%% file: main.tex
\RequirePackage[svgnames]{xcolor}

\documentclass[11pt,letterpaper]{mystyle}

\usepackage[all]{hypcap}
\usepackage[svgnames]{xcolor}
\usepackage[comma,authoryear,compress]{natbib}
\usepackage{hyperref}[citecolor=lightblue]

\hypersetup{
    colorlinks = true,
    citecolor = {YaleBlue},
}
\usepackage{enumitem}
\usepackage{algorithm}
\usepackage{algorithmicx}
\usepackage{algpseudocode}
\usepackage{microtype}
\usepackage{graphicx}
\expandafter\def\csname ver@subfig.sty\endcsname{}
\usepackage{booktabs} %
\usepackage{float}
\usepackage{bigstrut}

\usepackage{amsmath}
\usepackage{amssymb}
\usepackage{mathtools}
\usepackage{amsthm}
\usepackage{mathrsfs}
\usepackage{nicefrac}
\usepackage{dsfont}
\usepackage{enumitem}
\usepackage{subcaption}
\usepackage{graphicx,subfig}
\usepackage{cleveref}
\usepackage{bxcoloremoji}

\usepackage{float}

\usepackage[utf8]{inputenc} %
\usepackage[T1]{fontenc}    %
\usepackage{hyperref}       %
\usepackage{url}            %
\usepackage{booktabs}       %
\usepackage{amsfonts}       %
\usepackage{nicefrac}       %
\usepackage{microtype}      %
\usepackage{graphicx}
\usepackage{subcaption} %
\usepackage{amssymb}
\usepackage{fdsymbol}
\usepackage{wrapfig}
\usepackage{lipsum}
\usepackage{enumitem}
\usepackage{stackengine}
\usepackage[font=small,labelfont=bf]{caption}
\usepackage{color}
\usepackage{adjustbox}

\usepackage{rotating}
\usepackage{makecell}
\usepackage{url}
\usepackage{changepage}
\usepackage{graphicx}
\usepackage{makecell}
\usepackage{colortbl}
\usepackage{tikz}
\usepackage{amsmath}
\usepackage{xcolor}
\usepackage{tcolorbox}
\usepackage{multicol}
\usepackage{multirow}
\usepackage[utf8]{inputenc} %
\usepackage[T1]{fontenc}    %
\usepackage{hyperref}       %
\usepackage{url}            %
\usepackage{booktabs}       %
\usepackage{amsfonts}       %
\usepackage{nicefrac}       %
\usepackage{microtype}      %
\usepackage{graphicx}
\usepackage{wrapfig}
\usepackage{booktabs}
\usepackage{pifont} 
\usepackage{xspace}
\usepackage{subcaption}
\usepackage{paralist}

\input{macro}

\newtcolorbox{AIbox}[2][]{aibox,title=#2,#1}
\definecolor{lightblue}{rgb}{0.22,0.45,0.70}%
\definecolor{Gray}{gray}{0.95}
\definecolor{Cornsilk}{rgb}{1.0, 0.97, 0.86}

\usepackage{amsmath}

\usepackage[all]{hypcap}

\makeatletter
\renewcommand\paragraph{\@startsection{paragraph}{4}{\z@}%
	{0.7ex \@plus.2ex \@minus.2ex}%
	{-1em}%
	{\normalfont\normalsize\bfseries\maybe@addperiod}}
\newcommand{\maybe@addperiod}[1]{#1\@addpunct{.}}
\makeatother

\title{TAPS: Task Aware Proposal Distributions for Speculative Sampling}

\author{%
  Mohamad Zbib$^{1,2}$, Mohamad Bazzi$^{2}$, Ammar Mohanna$^{2}$, \\
  \textbf{Hasan Abed Al Kader Hammoud}$^{1,\ddag}$, \textbf{Bernard Ghanem}$^{1,\ddag}$ \\
  $^1$King Abdullah University of Science and Technology (KAUST) \quad
  $^2$American University of Beirut (AUB) \\
  $^\ddag$\texttt{Equal advising authors} \\
}

\runningtitle{TAPS: Task Aware Proposal Distributions for Speculative Sampling}

\correspondingauthor{
Mohamad Zbib \href{mailto:mbz02@mail.aub.edu}{mbz02@mail.aub.edu}; 
Hasan Abed Al Kader Hammoud \href{mailto:hasanabedalkader.hammoud@kaust.edu.sa}{hasanabedalkader.hammoud@kaust.edu.sa}
}
\begin{document}

\input{sections/abstract}

\maketitle
\vspace{3mm}
\input{sections/introduction}
\input{sections/relatedwork}
\input{sections/method}
\input{sections/experiments}
\input{sections/conclusion}
\clearpage 
\bibliography{main}

\appendix
\input{sections/appendix}
\appendix
\end{document}

%% file: macro.tex
\newcommand{\x}{\mathbf{x}}

\definecolor{blanchedalmond}{rgb}{1.0, 0.92, 0.8}
\definecolor{carmine}{rgb}{0.59, 0.0, 0.09}
\definecolor{lightblue}{rgb}{0.22,0.45,0.70}%

\renewcommand{\mathbf}{\boldsymbol}

\makeatletter
\def\Ddots{\mathinner{\mkern1mu\raise\p@
\vbox{\kern7\p@\hbox{.}}\mkern2mu
\raise4\p@\hbox{.}\mkern2mu\raise7\p@\hbox{.}\mkern1mu}}
\makeatother

\definecolor{amaranth}{rgb}{0.9, 0.17, 0.31}
\definecolor{antiquebrass}{rgb}{0.8, 0.58, 0.46}
\definecolor{antiquefuchsia}{rgb}{0.57, 0.36, 0.51}
\definecolor{chromeyellow}{rgb}{0.31, 0.47, 0.26}

\newcommand{\github}{\raisebox{-1.5pt}{\includegraphics[height=1.05em]{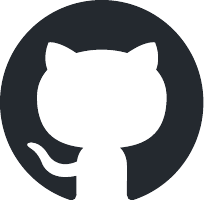}}}

%% file: sections/abstract.tex
\begin{abstract}
Speculative decoding accelerates autoregressive generation by letting a lightweight draft model propose future tokens that a larger target model then verifies in parallel. In practice, however, draft models are usually trained on broad generic corpora, which leaves it unclear how much speculative decoding quality depends on the draft training distribution. We study this question with lightweight HASS and EAGLE-2 drafters trained on MathInstruct, ShareGPT, and mixed-data variants, evaluated on MT-Bench, GSM8K, MATH-500, and SVAMP. Measured by acceptance length, task-specific training yields clear specialization: MathInstruct-trained drafts are strongest on reasoning benchmarks, while ShareGPT-trained drafts are strongest on MT-Bench. Mixed-data training improves robustness, but larger mixtures do not dominate across decoding temperatures. We also study how to combine specialized drafters at inference time. Naive checkpoint averaging performs poorly, whereas confidence-based routing improves over single-domain drafts and merged-tree verification yields the highest acceptance length overall for both backbones. Finally, confidence is a more useful routing signal than entropy: rejected tokens tend to have higher entropy, but confidence produces much clearer benchmark-level routing decisions. These results show that speculative decoding quality depends not only on draft architecture, but also on the match between draft training data and downstream workload, and that specialized drafters are better combined at inference time than in weight space.

\vspace{2mm}
\coloremojicode{1F4C5} \textbf{Date}: March 2026

\github{} \textbf{Code Repository}: 
\href{https://github.com/Moe-Zbeeb/TAPS}{https://github.com/Moe-Zbeeb/TAPS}

\coloremojicode{1F917} \textbf{Model Weights}: 
\href{https://huggingface.co/collections/zbeeb/taps}{https://huggingface.co/collections/zbeeb/taps}

\coloremojicode{1F4DA} \textbf{Datasets}: 
\href{https://huggingface.co/datasets/zbeeb/TAPS-Datasets}{https://huggingface.co/datasets/zbeeb/TAPS-Datasets}

\coloremojicode{1F4E7} \textbf{Contact}: 
\href{mailto:mbz02@mail.aub.edu}{mbz02@mail.aub.edu}, 
\href{mailto:hasanabedalkader.hammoud@kaust.edu.sa}{hasanabedalkader.hammoud@kaust.edu.sa}
\end{abstract}

%% file: sections/introduction.tex
\section*{Introduction}

Large language models (LLMs) achieve strong results across many tasks, but autoregressive decoding remains a major inference bottleneck because each token depends on the full previously generated prefix \cite{NEURIPS2020_1457c0d6, pmlr-v202-leviathan23a, chen2023acceleratinglargelanguagemodel}.

\begin{wrapfigure}{r}{0.35\textwidth}
\vspace{-12pt}
\centering
\includegraphics[width=\linewidth]{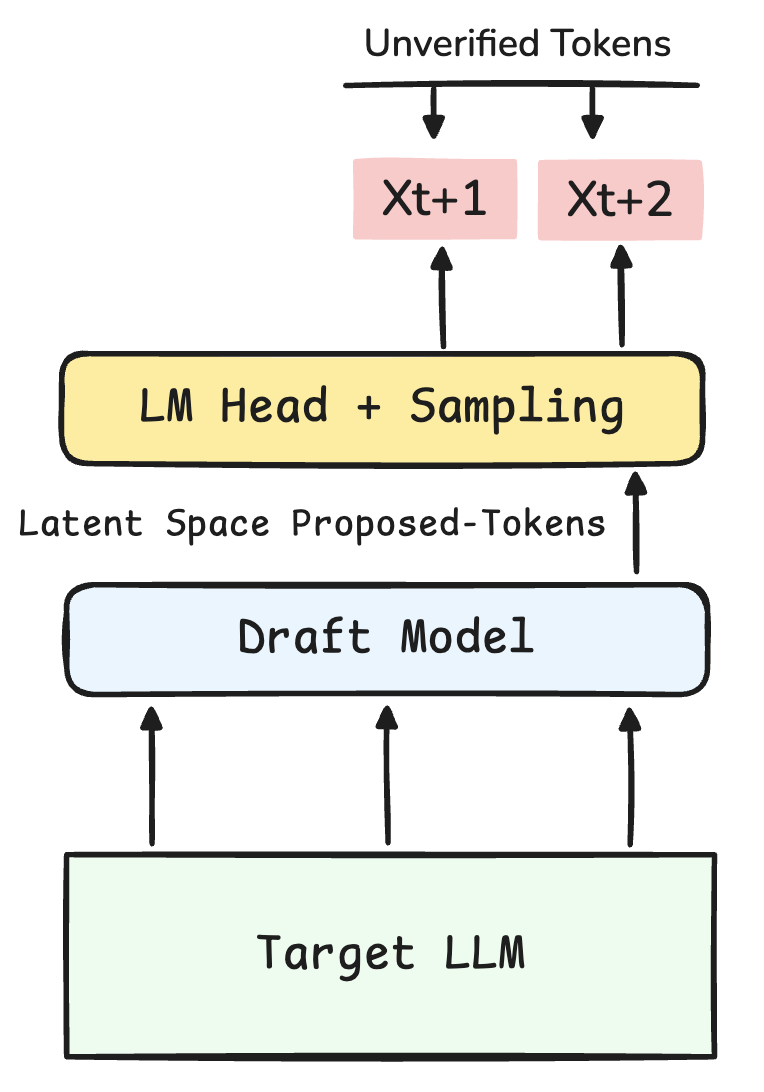}
\caption{Schematic of the speculative decoding pipeline. Given contextual information from the target LLM, the draft model generates latent proposed tokens, which are converted by the LM head and sampling module into multiple candidate future tokens. These candidates are provisional and are later verified by the target model. Importantly, the trainable component in this framework is the draft model, whose role is to efficiently approximate the target model’s next-token behavior while preserving the target model’s final output distribution after verification.}
\label{fig:illustrative}
\vspace{-8pt}
\end{wrapfigure}

Speculative decoding addresses this bottleneck by letting a lightweight draft model propose several future tokens that a larger target model then verifies in parallel. The appeal of the approach is that it can improve throughput without changing the target model's output distribution. Its usefulness, however, depends critically on the quality of the proposal distribution produced by the drafter. Figure~\ref{fig:illustrative} summarizes the setting studied in this paper.

Most prior work improves speculative decoding through better draft architectures or more efficient verification procedures. Early speculative decoding methods use a separate lightweight drafter \cite{pmlr-v202-leviathan23a, chen2023acceleratinglargelanguagemodel}, while later methods improve feature-level drafting and tree construction, as in EAGLE, EAGLE-2, EAGLE-3, and HASS \cite{EAGLE, EAGLE2, EAGLE3, zhang2025learning}. Other lines of work explore tree verification, self-speculative decoding, hierarchical drafting, cascaded drafters, and retrieval-assisted proposals \cite{Specinfer, zhang2024draft, elhoushi2024layerskip, liu2024kangaroo, sun2024triforce, chen2024cascade, he-etal-2024-rest}. Despite this progress, draft models are still typically trained on broad generic corpora such as ShareGPT, so the role of the draft training distribution remains under-studied.

This gap matters for two reasons. First, if the draft model is trained on a distribution that is poorly aligned with the target workload, acceptance length may degrade even when the speculative decoding algorithm itself is unchanged. Second, the open-weight ecosystem increasingly offers multiple specialized checkpoints \cite{sun2025survey}, which raises a practical question: when several specialized drafters are available, is it better to mix their data during training, merge them in weight space, or compose them at inference time \cite{ilharco2022editing, mu2026comprehensivesurveymixtureofexpertsalgorithms}?

We study these questions with lightweight drafters trained under HASS and EAGLE-2 on MathInstruct and ShareGPT. The paper is organized around five research questions. \textbf{RQ1.} Does task-specific training improve speculative decoding on matched downstream tasks? \textbf{RQ2.} Can mixed-data training recover cross-domain robustness without erasing specialization? \textbf{RQ3.} How should multiple specialized draft models be combined: weight averaging, routing, or merged-tree verification? \textbf{RQ4.} Are confidence, entropy, and depth-wise acceptance useful signals for explaining routing and acceptance behavior? \textbf{RQ5.} How does speculative depth affect the exploration and exploitation balance in task aware drafting?

Our answers are consistent across both speculative backbones. Single domain training produces clear specialization: MathInstruct-trained drafts are strongest on GSM8K, MATH-500, and SVAMP, while ShareGPT-trained drafts are strongest on MT-Bench. Mixed-data training improves robustness, but larger mixtures are not uniformly better across temperatures. When multiple specialists are available, naive weight-space averaging performs poorly, whereas inference-time composition is substantially stronger. Confidence-based routing improves over single-domain baselines, and merged-tree verification yields the highest acceptance length overall. Confidence is also more useful than entropy as a routing signal, although entropy remains informative as a diagnostic of likely rejection. Depth analysis also shows that speculative decoding becomes increasingly dominated by the task matched specialist at larger depths. The rest of the paper follows the same logic: we first review the decoding backbones, then describe the controlled setup, and finally answer the five research questions with matched evidence.

\begin{tcolorbox}[colback=Cornsilk,colframe=white,boxrule=0pt,sharp corners,left=6pt,right=6pt,top=4pt,bottom=4pt]
\footnotesize \textit{All training and evaluation experiments in this paper were run on a single node with four NVIDIA A100 GPUs.}
\end{tcolorbox}

%% file: sections/relatedwork.tex
\section*{Preliminaries}

We briefly review only the pieces of speculative decoding that are needed later: the lossless verification rule and the two drafting backbones used in our experiments, EAGLE-2 and HASS. Throughout the paper, the verifier and acceptance rule are fixed; what changes is the draft training distribution and the way multiple specialized drafts are composed at test time.

\subsection*{Speculative Decoding}

Speculative decoding uses a lightweight draft model $p$ to propose $K$ future tokens and a target model $q$ to verify them in parallel, thereby reducing the number of expensive target-model calls. Given a prefix $x_{1:n}$, the draft model generates $\tilde{x}_{n+1:n+K}$ autoregressively. The target model then scores these candidates, and each drafted token $\tilde{x}_{n+t}$ is accepted sequentially with probability
\begin{equation}
\alpha_{n+t}
=
\min\!\left(
1,\;
\frac{q(\tilde{x}_{n+t}\mid x_{1:n+t-1})}{p(\tilde{x}_{n+t}\mid x_{1:n+t-1})}
\right).
\end{equation}
If rejection occurs, decoding instead samples from
\begin{equation}
r(x)
\propto
\max\!\left(0,\; q(x\mid x_{1:n+t-1}) - p(x\mid x_{1:n+t-1})\right).
\end{equation}
This rejection-sampling correction preserves the target model's output distribution exactly while allowing multiple proposed tokens to be checked in a single verifier call. In the rest of the paper, acceptance length measures how often this lossless verification procedure approves long draft continuations.

\subsection*{EAGLE-2}

EAGLE-2 inherits the EAGLE feature-level drafter and improves inference by replacing a fixed draft tree with a context-dependent dynamic tree \cite{EAGLE2}. Let $h_t$ denote the target model's second-to-last-layer feature at step $t$. Instead of autoregressing directly on tokens, the draft model $g_\phi$ predicts future features,
\begin{equation}
\hat{h}_{t+1} = g_\phi(\hat{h}_{\le t}, x_{\le t+1}),
\end{equation}
and the predicted features are mapped to token probabilities through the target model's LM head. A compact training objective is
\begin{equation}
\mathcal{L}_{\text{EAGLE}}
=
\sum_t \|\hat{h}_{t+1}-h_{t+1}\|_2^2
+
\lambda \sum_t \mathrm{CE}\!\left(\mathrm{softmax}(W\hat{h}_{t+1}), x_{t+1}\right).
\end{equation}
During verification, a drafted token $\hat{x}_{j+i}$ is accepted with probability
\begin{equation}
\alpha_{j+i}
=
\min\!\left(1,\frac{p_{j+i}(\hat{x}_{j+i})}{\hat{p}_{j+i}(\hat{x}_{j+i})}\right),
\end{equation}
which preserves the target model's output distribution. EAGLE-2 uses draft confidence to rank frontier nodes in a dynamic draft tree,
\begin{equation}
V_i \approx \prod_{v_j \in \mathrm{Path}(\mathrm{root}, v_i)} c_j,
\end{equation}
and expands the highest-valued frontier nodes before verification \cite{EAGLE2}. In this paper, we keep that decoding rule fixed and vary only how the drafter is trained or combined.

\subsection*{HASS}

HASS uses the same lossless speculative decoding framework, but improves the drafter by reducing objective mismatch and context mismatch between training and inference \cite{zhang2025learning}. Its harmonized objective distillation term focuses learning on the verifier's most likely next tokens. Let $q(\cdot)$ and $p(\cdot)$ denote the next-token distributions of the target and draft models, respectively, and let $\hat{\Omega}\subset\Omega$ be the set of top-$K$ tokens under $q$. The Top-$K$ distillation loss is
\begin{equation}
\mathcal{L}_{\mathrm{Top\text{-}K}}
=
-\sum_{x \in \hat{\Omega}} q(x)\log p(x).
\end{equation}
HASS also introduces harmonized context alignment so that later draft predictions are trained on imperfect draft features rather than only clean target features. At alignment step $j$, the draft model predicts
\begin{equation}
P^{(s)}(x_{t+1}\mid x_{\le t})
=
\mathrm{Head}\!\left(f^{(s_j)}_{t+1}\right)
=
\mathrm{Head}\!\left(
M^{(s)}\!\left(
f^{(s_{j-1})}_t,\;
f^{(l)}_{1}\oplus\cdots\oplus f^{(l)}_{t-j+1}
\oplus
f^{(s_1)}_{t-j+2}\oplus\cdots\oplus f^{(s_{j-1})}_{t}
\right)
\right),
\end{equation}
with training objective
\begin{equation}
\mathcal{L}_{\mathrm{HASS}}^{(j)}
=
\sum_{t=1}^{T-1}
\Big[
\mathrm{CE}\!\big(
P^{(l)}(x_{t+1}\mid x_{\le t}),
P^{(s)}(x_{t+1}\mid x_{\le t})
\big)
+
\mathcal{L}_{\mathrm{aux}}
\Big],
\end{equation}
where $\mathcal{L}_{\mathrm{aux}}$ includes Top-$K$ distillation and feature regression. As with EAGLE-2, our experiments keep the verifier and the lossless acceptance rule fixed and study how different training distributions and composition strategies affect acceptance.

%% file: sections/method.tex
\section*{Experimental Setup and Composition Strategies}

\subsection*{Common Setup}

We study task-aware draft construction for speculative decoding under a fixed verifier. Across all experiments, the verifier is Meta-Llama-3-8B-Instruct \cite{grattafiori2024llama3herdmodels}, and the draft model is a lightweight LLaMA-style decoder with one transformer layer, hidden size 4096, and roughly 0.8B parameters. Draft and target share the same tokenizer and vocabulary so that acceptance differences are not confounded by tokenization mismatch.

We evaluate on MT-Bench together with three reasoning-heavy benchmarks, GSM8K, MATH-500, and SVAMP, at temperatures 0 and 1. Our primary metric is acceptance length averaged over the evaluation distribution under the lossless speculative decoding constraint. Unless stated otherwise, the only factors that change across experiments are the draft training distribution or the way multiple specialized drafts are combined.

All draft checkpoints are trained for 20 epochs with learning rate $3 \times 10^{-5}$, batch size 8, and gradient accumulation 1. HASS runs use the same auxiliary settings throughout: top-$K$ distillation with $K=10$, loss weight 1.0, and three forward-alignment steps. The study varies along four axes: speculative backbone (EAGLE-2 or HASS), training domain (ShareGPT for conversational data or MathInstruct for mathematical reasoning), mixed-data supervision (35k+35k or 70k+70k), and test-time composition strategy (Averaged, Confidence Routed, or Merged Trees).

For an input $x$, verifier $M_T$, and drafter $M_D$, let $A(x; M_D, M_T)$ denote the number of consecutively accepted draft tokens. We compare methods through
\begin{equation}
\mathbb{E}_{x \sim \mathcal{D}}[A(x; M_D, M_T)],
\end{equation}
subject to the lossless speculative decoding constraint that the final output distribution remains identical to that of the verifier.

Unless stated otherwise, the only factors that change across experiments are the draft training data or the way multiple specialized drafts are combined.

\subsection*{Draft Variants}

We study seven main draft variants for each backbone. Two are single-domain checkpoints: one trained on 70k MathInstruct examples and one trained on 70k ShareGPT examples. Two are mixed-data checkpoints: Mixed 35k+35k and Mixed 70k+70k. The remaining three use the single-domain checkpoints as building blocks for composition: Averaged, Confidence Routed, and Merged Trees.

These variants map directly to the research questions in the experiments section. The single-domain checkpoints answer RQ1, the mixed-data checkpoints answer RQ2, and the three composition strategies answer RQ3. RQ4 then interprets these results through routing statistics, entropy, and depth-wise acceptance.

\subsection*{Checkpoint Averaging}
\begin{figure*}[t]
\centering

\begin{subfigure}[t]{0.49\textwidth}
    \centering
    \includegraphics[width=\linewidth]{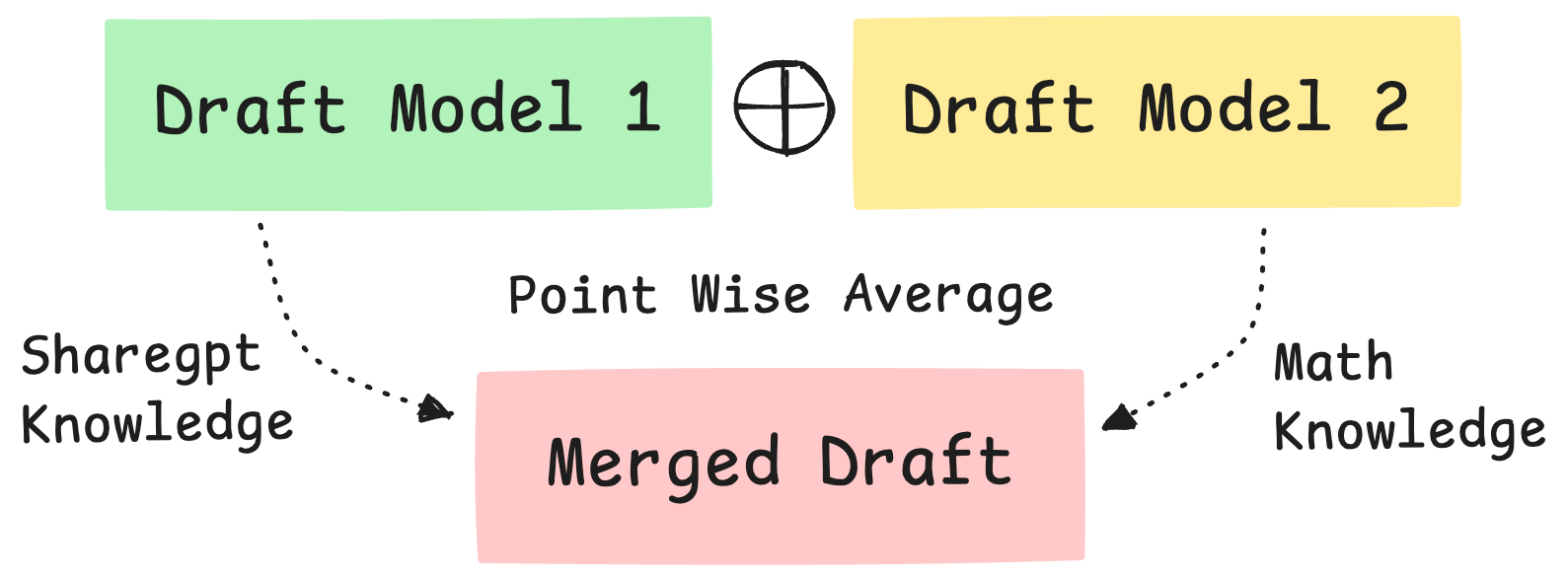}
    \caption{Checkpoint weight merging by point-wise parameter averaging. Draft models trained on ShareGPT and MathInstruct are combined by averaging corresponding parameters in weight space, producing a single merged draft that retains information from both domains.}
    \label{fig:merge}
\end{subfigure}
\hfill
\begin{subfigure}[t]{0.39\textwidth}
    \centering
    \includegraphics[width=\linewidth]{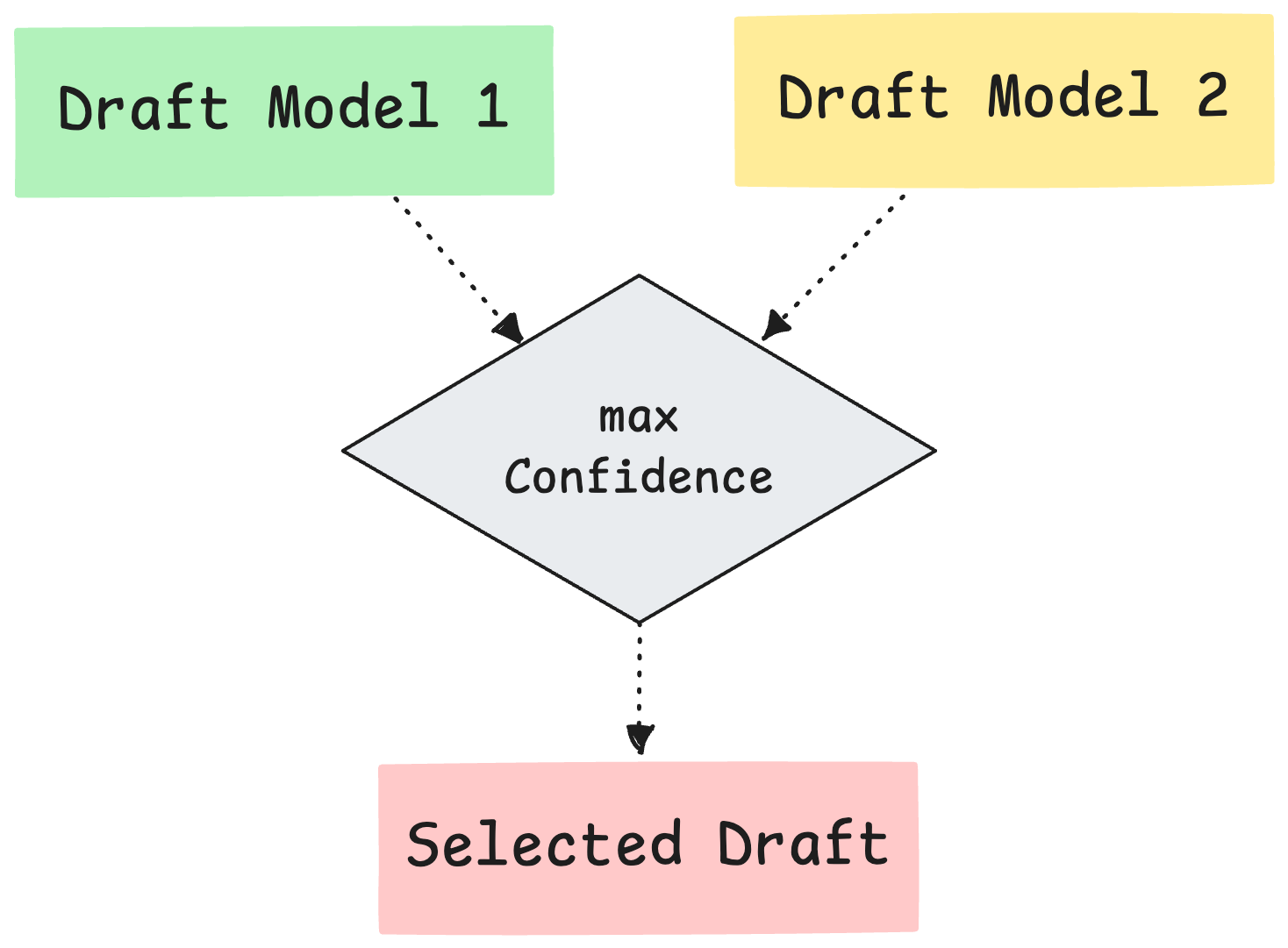}
    \caption{Confidence-based routing between specialized draft models. At inference time, the system selects the draft with the highest confidence for the current prompt, enabling task-aware use of specialized models without merging their parameters.}
    \label{fig:routing}
\end{subfigure}

\caption{Two strategies for combining specialized draft models. Left: checkpoint weight merging in parameter space. Right: confidence-based routing at inference time.}
\label{fig:merge_routing}
\vspace{-8pt}
\end{figure*}

Our simplest composition baseline is checkpoint averaging. Let $\theta_{\mathrm{math}}$ and $\theta_{\mathrm{chat}}$ denote the parameters of the MathInstruct and ShareGPT draft models. We define the merged checkpoint as
\begin{equation}
\theta_{\mathrm{merge}}
=
\lambda \theta_{\mathrm{math}} + (1-\lambda)\theta_{\mathrm{chat}},
\end{equation}
where $\lambda \in [0,1]$ controls the contribution of each checkpoint. We use $\lambda = 0.5$ for the main table and sweep $\lambda$ in Figure~\ref{fig:averaging}.

\subsection*{Inference-Time Composition}

We also study two inference-time alternatives that keep the single-domain checkpoints separate. Both strategies operate within one backbone at a time: HASS drafts are combined only with HASS drafts, and EAGLE-2 drafts only with EAGLE-2 drafts. This isolates the effect of composition from any cross-backbone differences.

\paragraph{Confidence routing} Given an input prefix, we decode one draft tree from the MathInstruct checkpoint and one from the ShareGPT checkpoint. We score each tree by its mean draft confidence. Let $\mathcal{T}_{\mathrm{math}}$ and $\mathcal{T}_{\mathrm{chat}}$ denote the two trees and let $c(v)$ denote the confidence assigned to node $v$. The tree-level score is
\begin{equation}
\mathrm{Score}(\mathcal{T})
=
\frac{1}{|\mathcal{T}|}\sum_{v \in \mathcal{T}} c(v),
\end{equation}
and the selected tree is
\begin{equation}
\mathcal{T}^{*}
=
\arg\max_{\mathcal{T} \in \{\mathcal{T}_{\mathrm{math}}, \mathcal{T}_{\mathrm{chat}}\}}
\mathrm{Score}(\mathcal{T}).
\end{equation}
Only $\mathcal{T}^{*}$ is passed to the verifier.

\begin{figure*}[!t]
\centering
\includegraphics[width=\textwidth]{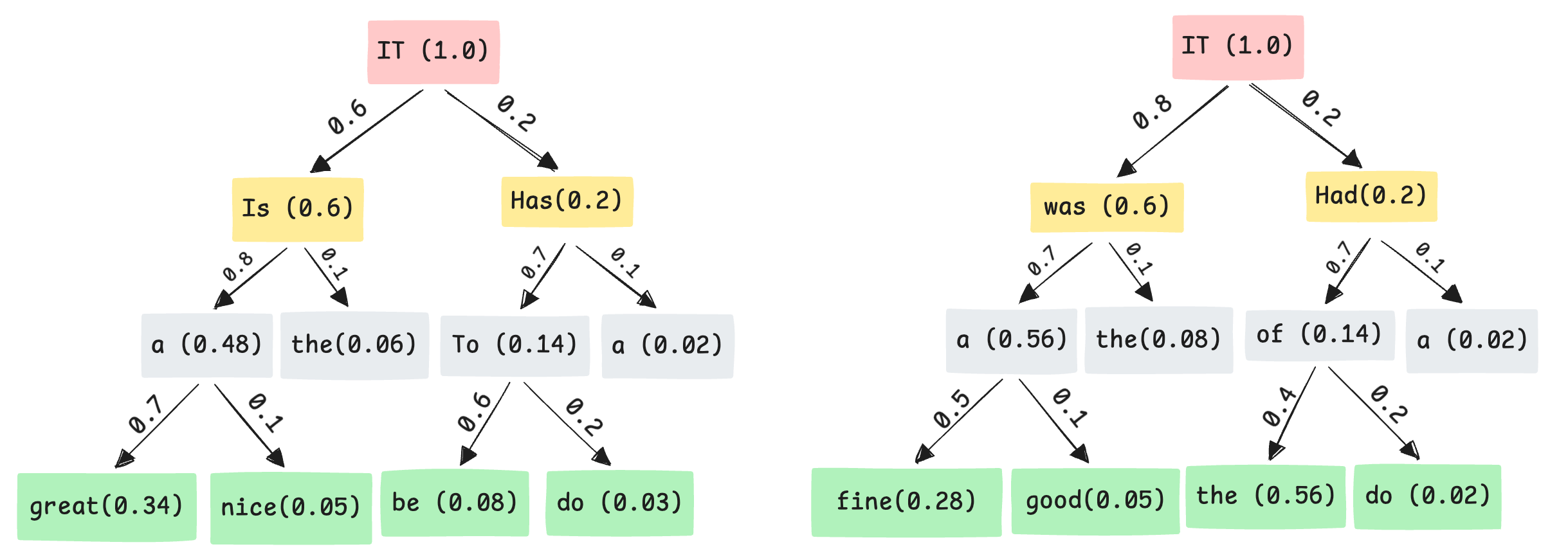}
\caption{\textbf{Confidence Routing Between Specialized Trees.} The MathInstruct and ShareGPT checkpoints generate separate draft trees from the same prefix, with node labels indicating draft confidence. Confidence routing selects the tree with the higher mean node confidence before verification.}
\label{fig:trees}
\end{figure*}

\paragraph{Merged-tree verification} Instead of selecting one tree and discarding the other, we can verify both trees jointly by packing them under a shared root. We preserve the node indices of one subtree, offset the other subtree by the size of the first, and build an attention mask that allows each node to attend only to the shared root and its own ancestors. Candidates from the MathInstruct subtree therefore do not attend to candidates from the ShareGPT subtree, and vice versa. Position ids are assigned by tree depth so that each child is placed one step deeper than its parent.

The merged tree increases proposal diversity at each verifier call because both specialists contribute candidate continuations. At the same time, it is a stricter test than routing because the verifier must process a larger tree. In this paper we report the acceptance length benefit of this strategy, but we do not claim an end-to-end latency improvement without a separate systems analysis.

\begin{figure*}[!t]
\centering
\begin{minipage}[t]{0.42\textwidth}
\captionof{algorithm}{\textbf{Merged-Tree Verification.}}
\label{alg:merged_tree_verification}
\small
\begin{algorithmic}[1]
    \Require Prefix $y_{1:t}$, target $M_T$, drafts $M_{\mathrm{math}}, M_{\mathrm{chat}}$
\State Generate draft trees $\mathcal{T}_{\mathrm{math}}$ and $\mathcal{T}_{\mathrm{chat}}$ from the same root token
\State Merge the two trees under a shared root by concatenating nodes and remapping indices
\State Build ancestor-preserving attention masks and depth-based position ids
\State Verify the merged tree in one parallel pass with $M_T$
\State Extract candidate paths and apply standard speculative acceptance
\State Commit the accepted prefix
\State \Return accepted length
\end{algorithmic}
\end{minipage}
\hfill
\begin{minipage}[t]{0.52\textwidth}
\vspace{-4pt}
\centering
\includegraphics[width=\linewidth]{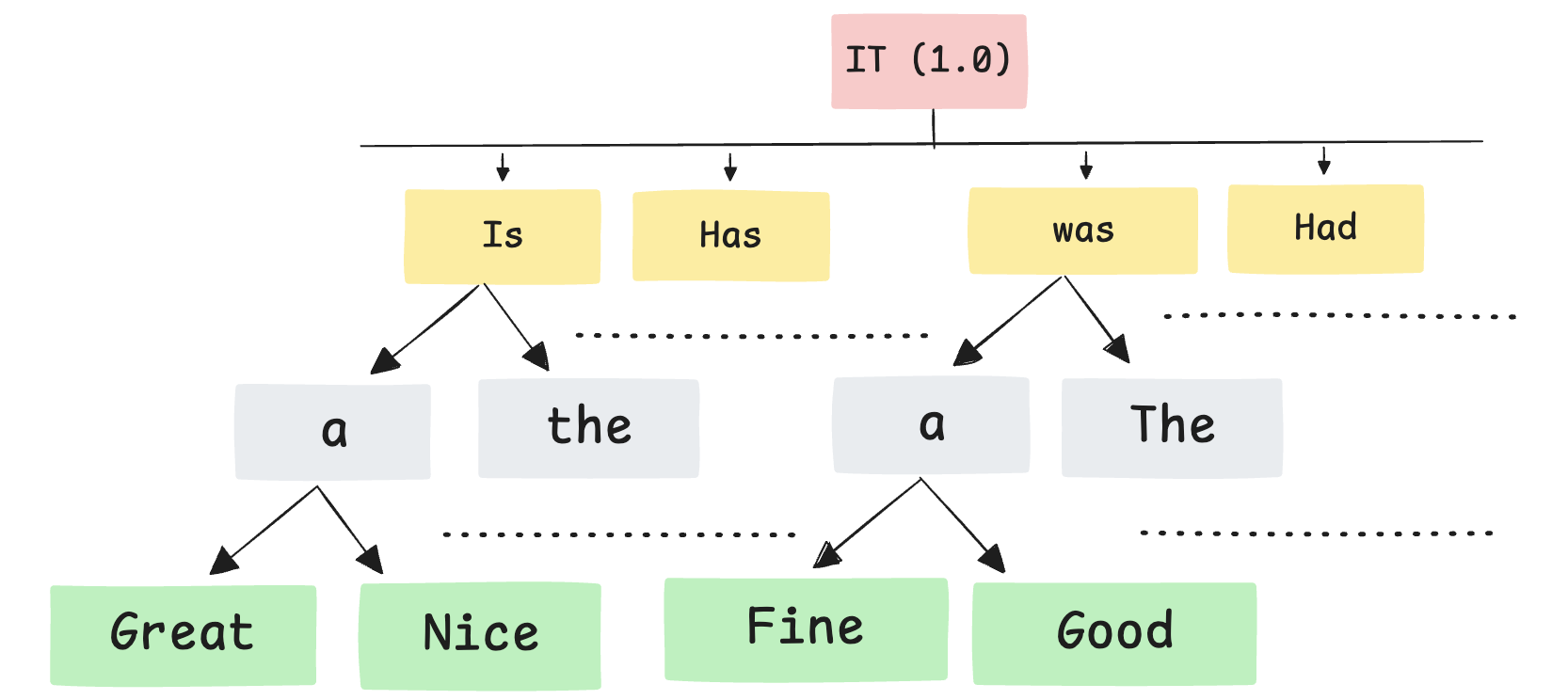}
\captionof{figure}{\textbf{Merged Verification Tree.} The MathInstruct and ShareGPT subtrees are packed under a shared root while preserving their internal ancestry. This lets the verifier evaluate both specialists in one pass and tests whether broader proposal coverage is more useful than selecting a single specialist.}
\label{fig:merged_trees}
\end{minipage}
\end{figure*}

\begin{figure}[H]
\centering
\includegraphics[width=\linewidth]{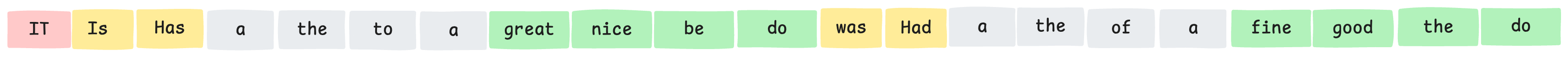}
\caption{\textbf{Flattened Merged-Tree Input.} The merged tree is serialized for verification while ancestry is preserved through the tree attention mask and depth-based position ids. This lets the verifier process both specialized subtrees without cross-subtree attention.}
\label{fig:flattened_merged_tree}
\end{figure}

%% file: sections/experiments.tex
\section*{Experiments}

We report acceptance length, the average number of draft tokens accepted per verifier call. Higher acceptance length indicates that the drafter is better aligned with the verifier on the evaluated workload. Table~\ref{tab:main_results} reports the main numbers. The section then reads those results in four passes: single-domain specialization (RQ1), mixed-data robustness (RQ2), specialist composition (RQ3), and routing/diagnostic signals (RQ4). Unless stated otherwise, all conclusions below are drawn from Table~\ref{tab:main_results}, Table~\ref{tab:routing}, and the supporting figures.

\begin{table}[htb]
\centering
\caption{\textbf{Main Results by Research Question.} Average acceptance length on MT-Bench, GSM8K, MATH-500, and SVAMP for HASS and EAGLE-2 at temperatures $0$ and $1$. Rows are grouped by the question they answer: RQ1 tests single-domain specialization, RQ2 mixed-data robustness, and RQ3 composition strategies. Higher is better.}
\label{tab:main_results}
\resizebox{\textwidth}{!}{
\begingroup
\arrayrulecolor{black}
\rowcolors{3}{Cornsilk}{white}
\renewcommand{\arraystretch}{1.1}
\begin{tabular}{l l c c c c c c c c c c}
\toprule
 & & \multicolumn{5}{c}{Temperature 0} & \multicolumn{5}{c}{Temperature 1} \\
Model Variant & Method & MT-Bench & GSM8K & MATH-500 & SVAMP & Average & MT-Bench & GSM8K & MATH-500 & SVAMP & Average \\
\midrule
\rowcolor{white}
\multicolumn{12}{l}{\textbf{RQ1. Task-specific training: single-domain checkpoints}} \\
MathInstruct & HASS & 2.90 & 5.02 & 5.35 & 3.13 & 4.10 & 2.31 & 4.75 & 4.63 & 2.46 & 3.54 \\
MathInstruct & EAGLE-2 & 2.54 & 5.04 & 5.28 & 4.81 & 4.42 & 2.43 & 4.71 & 4.61 & 4.53 & 4.07 \\
ShareGPT & HASS & 3.98 & 4.09 & 3.98 & 4.44 & 4.12 & 3.50 & 4.03 & 3.61 & 3.95 & 3.77 \\
ShareGPT & EAGLE-2 & 3.57 & 3.72 & 3.81 & 3.71 & 3.70 & 3.38 & 3.72 & 3.43 & 3.65 & 3.54 \\
\midrule
\rowcolor{white}
\multicolumn{12}{l}{\textbf{RQ2. Mixed-data training: robustness checkpoints}} \\
Mixed 35k+35k & HASS & 3.92 & 4.77 & 5.02 & 4.15 & 4.47 & 3.46 & 4.66 & 4.47 & 4.57 & 4.29 \\
Mixed 35k+35k & EAGLE-2 & 3.37 & 4.12 & 4.44 & 4.16 & 4.02 & 3.10 & 4.08 & 4.02 & 4.03 & 3.81 \\
Mixed 70k+70k & HASS & 4.13 & 5.53 & 5.67 & 5.38 & 5.18 & 3.17 & 4.16 & 3.42 & 4.01 & 3.69 \\
Mixed 70k+70k & EAGLE-2 & 3.75 & 4.68 & 4.85 & 4.64 & 4.48 & 2.99 & 3.76 & 3.20 & 3.08 & 3.26 \\
\midrule
\rowcolor{white}
\multicolumn{12}{l}{\textbf{RQ3. Combining specialists: weight averaging vs. inference-time composition}} \\
Averaged & HASS & 2.29 & 2.80 & 3.12 & 2.13 & 2.59 & 2.10 & 2.78 & 2.90 & 2.69 & 2.62 \\
Averaged & EAGLE-2 & 2.07 & 2.53 & 2.57 & 2.50 & 2.42 & 2.01 & 2.49 & 2.42 & 2.45 & 2.34 \\
Confidence Routed & HASS & 3.93 & 5.01 & 5.37 & 4.89 & 4.80 & 3.51 & 4.72 & 4.55 & 4.71 & 4.37 \\
Confidence Routed & EAGLE-2 & 3.63 & 4.91 & 5.25 & 4.71 & 4.63 & 3.36 & 4.65 & 4.62 & 4.46 & 4.27 \\
Merged Trees & HASS & 4.05 & 5.42 & 5.65 & 5.31 & 5.11 & 3.76 & 5.21 & 4.98 & 5.05 & 4.75 \\
Merged Trees & EAGLE-2 & 3.93 & 5.32 & 5.63 & 5.25 & 5.03 & 3.55 & 5.01 & 4.79 & 4.93 & 4.57 \\
\bottomrule
\end{tabular}
\endgroup
}
\end{table}

Each subsection reads the relevant block of Table~\ref{tab:main_results}, points to the supporting figures when needed, and states the narrowest conclusion supported by that evidence.

\subsection*{RQ1: Does task-specific training improve matched-domain acceptance?}

\paragraph{Question} Do drafters trained on a matched domain achieve longer acceptance lengths than drafters trained on a mismatched domain?

\paragraph{Setup} We compare the two single-domain checkpoints in Table~\ref{tab:main_results}. MathInstruct is intended to specialize on mathematical reasoning, while ShareGPT is intended to specialize on conversational generation. Figure~\ref{fig:depths} provides a depth-wise view of the same behavior.

\paragraph{Answer} Yes. The specialization pattern is clear at temperature $0$ for both backbones. Under HASS, ShareGPT is stronger than MathInstruct on MT-Bench (3.98 vs.\ 2.90), while MathInstruct is stronger on GSM8K and MATH-500 (5.02 vs.\ 4.09 and 5.35 vs.\ 3.98). The same pattern appears under EAGLE-2, where ShareGPT is strongest on MT-Bench (3.57 vs.\ 2.54) and MathInstruct is strongest on GSM8K, MATH-500, and SVAMP. Figure~\ref{fig:depths} shows that the specialization persists across speculative depth, especially on reasoning-heavy tasks.

\paragraph{Takeaway} Draft quality is not only a property of the speculative decoding backbone. It also depends on whether the draft training distribution matches the downstream workload.

\subsection*{RQ2: Can mixed-data training recover cross-domain robustness?}

\paragraph{Question} If single-domain drafts specialize strongly, can mixed-data training produce a more robust single checkpoint?

\paragraph{Setup} We compare the two mixed-data checkpoints against the single-domain checkpoints in Table~\ref{tab:main_results}. The Mixed 35k+35k variant keeps the total sample count modest and balanced, while Mixed 70k+70k doubles the amount of mixed supervision.

\paragraph{Answer} Mixed-data training improves robustness, but the effect is not monotonic. Under HASS at temperature $0$, Mixed 70k+70k is the strongest trained checkpoint overall with average acceptance length 5.18, but at temperature $1$ it falls to 3.69, below Mixed 35k+35k at 4.29. Under EAGLE-2, the same pattern appears: Mixed 70k+70k is strongest at temperature $0$ (4.48), whereas Mixed 35k+35k is more stable at temperature $1$ (3.81 vs.\ 3.26). Mixed training broadens coverage, but larger mixtures do not uniformly improve generalization across decoding temperatures.

\paragraph{Takeaway} Mixed-data training is a useful robustness strategy, but it does not remove the need to tune the mixture for the decoding regime of interest.

\subsection*{RQ3: How should multiple specialized drafters be combined?}

\paragraph{Question} When multiple specialized drafts are available, is it better to merge them in weight space or compose them at inference time?

\begin{figure}[!t]
    \centering
    \includegraphics[width=1\linewidth]{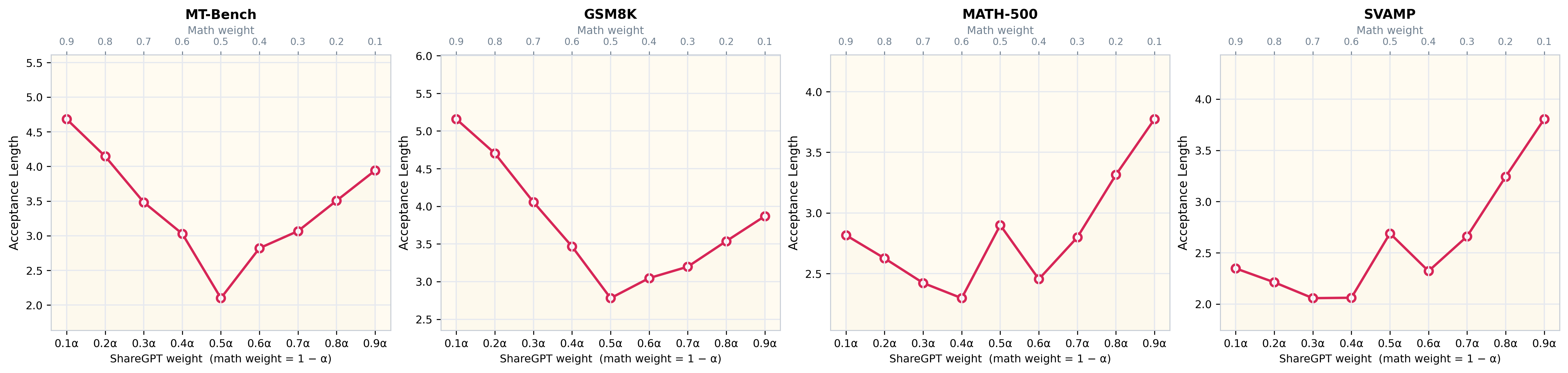}
    \caption{\textbf{Interpolation Sweep for Checkpoint Averaging.} Acceptance length is plotted against the interpolation weight between the MathInstruct and ShareGPT draft checkpoints under a fixed verifier setup. Weight-space averaging is unstable and remains well below the strongest inference-time composition methods.}
    \label{fig:averaging}
\end{figure}

\paragraph{Setup} We compare checkpoint averaging, confidence routing, and merged-tree verification in Table~\ref{tab:main_results}. Figure~\ref{fig:averaging} provides the interpolation sweep for checkpoint averaging.

\paragraph{Answer} Inference-time composition is substantially stronger than weight-space averaging. Averaged checkpoints are consistently the weakest variants in the main table, with average acceptance length between 2.34 and 2.62 across methods and temperatures. By contrast, confidence routing improves to 4.80 and 4.63 average acceptance length at temperature $0$ under HASS and EAGLE-2, respectively. Merged-tree verification is strongest overall, reaching 5.11 for HASS and 5.02 for EAGLE-2 at temperature $0$, and remaining the best variant at temperature $1$ as well. Figure~\ref{fig:averaging} reinforces this result: interpolating between the two checkpoints produces unstable behavior and never approaches the best inference-time composition methods.

\paragraph{Takeaway} If multiple specialists are available, they should be kept separate and combined at inference time. Weight-space averaging is a weak baseline for this problem.

\subsection*{RQ4: What do confidence, entropy, and depth reveal about acceptance behavior?}

\paragraph{Question} Are confidence and entropy useful signals for routing, and does depth-wise acceptance help explain the observed specialization?

\begin{figure}[!t]
    \centering
    \begin{subfigure}{\linewidth}
        \centering
        \includegraphics[width=1\linewidth]{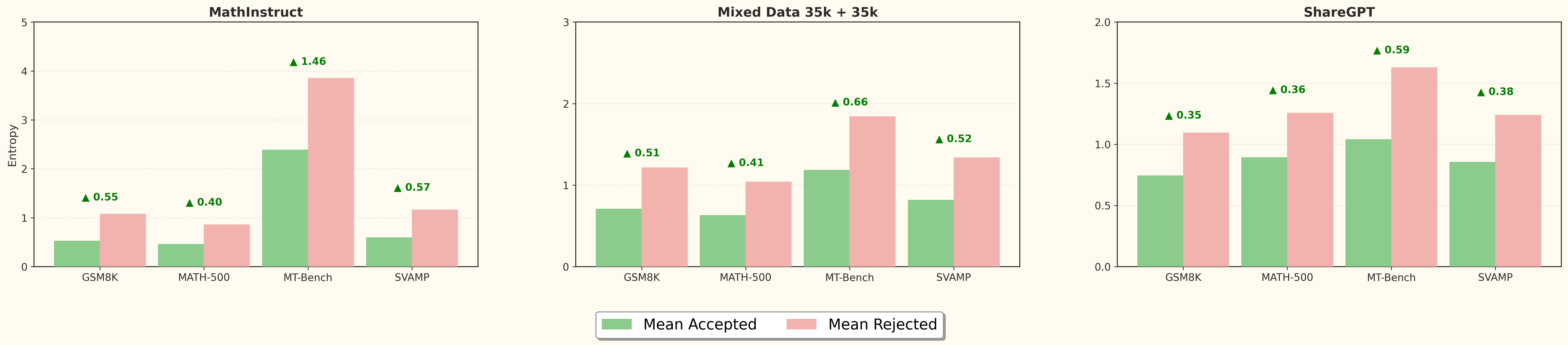}
        \caption{\textbf{EAGLE-2 Entropy.} Draft entropy for accepted and rejected tokens at temperature $0$ across benchmarks and checkpoints. The averaged checkpoint is omitted for readability. Rejected tokens consistently exhibit higher entropy.}
        \label{fig:eagle2_draft_entropy}
    \end{subfigure}

    \vspace{0.8em}

    \begin{subfigure}{\linewidth}
        \centering
        \includegraphics[width=1\linewidth]{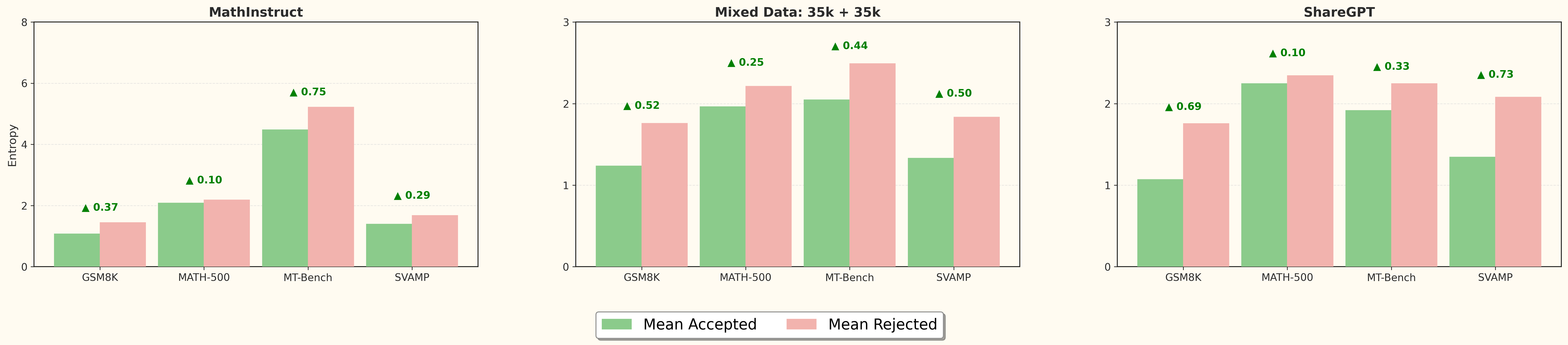}
        \caption{\textbf{HASS Entropy.} Draft entropy for accepted and rejected tokens at temperature $0$ across benchmarks and checkpoints. The averaged checkpoint is omitted for readability. The same accepted-versus-rejected separation remains visible.}
        \label{fig:hass_draft_entropy}
    \end{subfigure}

\caption{\textbf{Accepted vs. Rejected Token Entropy.} Both panels compare draft entropy at temperature $0$ for EAGLE-2 and HASS on the same benchmark suite and checkpoint families. Entropy is a useful diagnostic of rejection, but Table~\ref{tab:routing} shows that it is weaker than confidence for routing.}
\label{fig:draft_entropy_combined}
\end{figure}

\paragraph{Setup} Table~\ref{tab:routing} compares benchmark-level routing decisions under confidence-based and entropy-based selection for EAGLE-2. Figure~\ref{fig:draft_entropy_combined} compares accepted and rejected-token entropy, and Figure~\ref{fig:depths} reports acceptance by speculative depth for all main variants.

\paragraph{Answer} Confidence is useful for routing; entropy is mainly diagnostic. Under confidence routing, the MathInstruct drafter is selected for 90.8\% of GSM8K, 97.0\% of MATH-500, and 93.0\% of SVAMP examples, while ShareGPT is selected for 81.2\% of MT-Bench examples. Entropy routing is far less discriminative, producing near-balanced splits across all benchmarks. The entropy figures still show a consistent descriptive pattern: rejected tokens tend to have higher entropy than accepted tokens for both HASS and EAGLE-2. Figure~\ref{fig:depths} adds a second clue: acceptance falls with speculative depth for every variant, but domain specialization remains visible and often becomes more pronounced deeper in the tree.

\paragraph{Takeaway} Confidence is the stronger decision signal for routing between specialized drafters. Entropy and depth-wise acceptance are useful for interpreting verifier failures, but they do not by themselves justify a stronger routing policy.

\begin{table}[H]
\centering
\tiny
\resizebox{0.92\textwidth}{!}{
\begingroup
\setlength{\tabcolsep}{4pt}
\arrayrulecolor{black}
\rowcolors{3}{Cornsilk}{white}
\renewcommand{\arraystretch}{1.1}
\begin{tabular}{l c c c c c c}
\toprule
& \multicolumn{3}{c}{Confidence Routing} & \multicolumn{3}{c}{Entropy Routing} \\
\cmidrule(lr){2-4} \cmidrule(lr){5-7}
Benchmark & MathInstruct & ShareGPT & Total & MathInstruct & ShareGPT & Total \\
\midrule
MT-Bench & 15 (18.8\%) & 65 (81.2\%) & 80 & 42 (52.5\%) & 38 (47.5\%) & 80 \\
GSM8K & 1198 (90.8\%) & 121 (9.2\%) & 1319 & 720 (54.6\%) & 599 (45.4\%) & 1319 \\
MATH-500 & 485 (97.0\%) & 15 (3.0\%) & 500 & 312 (62.4\%) & 188 (37.6\%) & 500 \\
SVAMP & 279 (93.0\%) & 21 (7.0\%) & 300 & 159 (53.0\%) & 141 (47.0\%) & 300 \\
\bottomrule
\end{tabular}
\endgroup
}
\normalsize
    \caption{\textbf{Routing Decisions by Benchmark.} Benchmark-level routing counts for EAGLE-2 under confidence-based and entropy-based selection. Confidence routing separates conversational and mathematical workloads much more clearly than entropy routing.}
\label{tab:routing}
\end{table}

\subsection*{RQ5: How does speculative depth affect the exploration and exploitation balance in task aware drafting?}

\paragraph{Question} Does speculative depth reveal a shift from broad proposal coverage at shallow levels to stronger reliance on a task matched specialist at deeper levels?

\begin{figure}[H]
    \centering
    \includegraphics[width=1\linewidth]{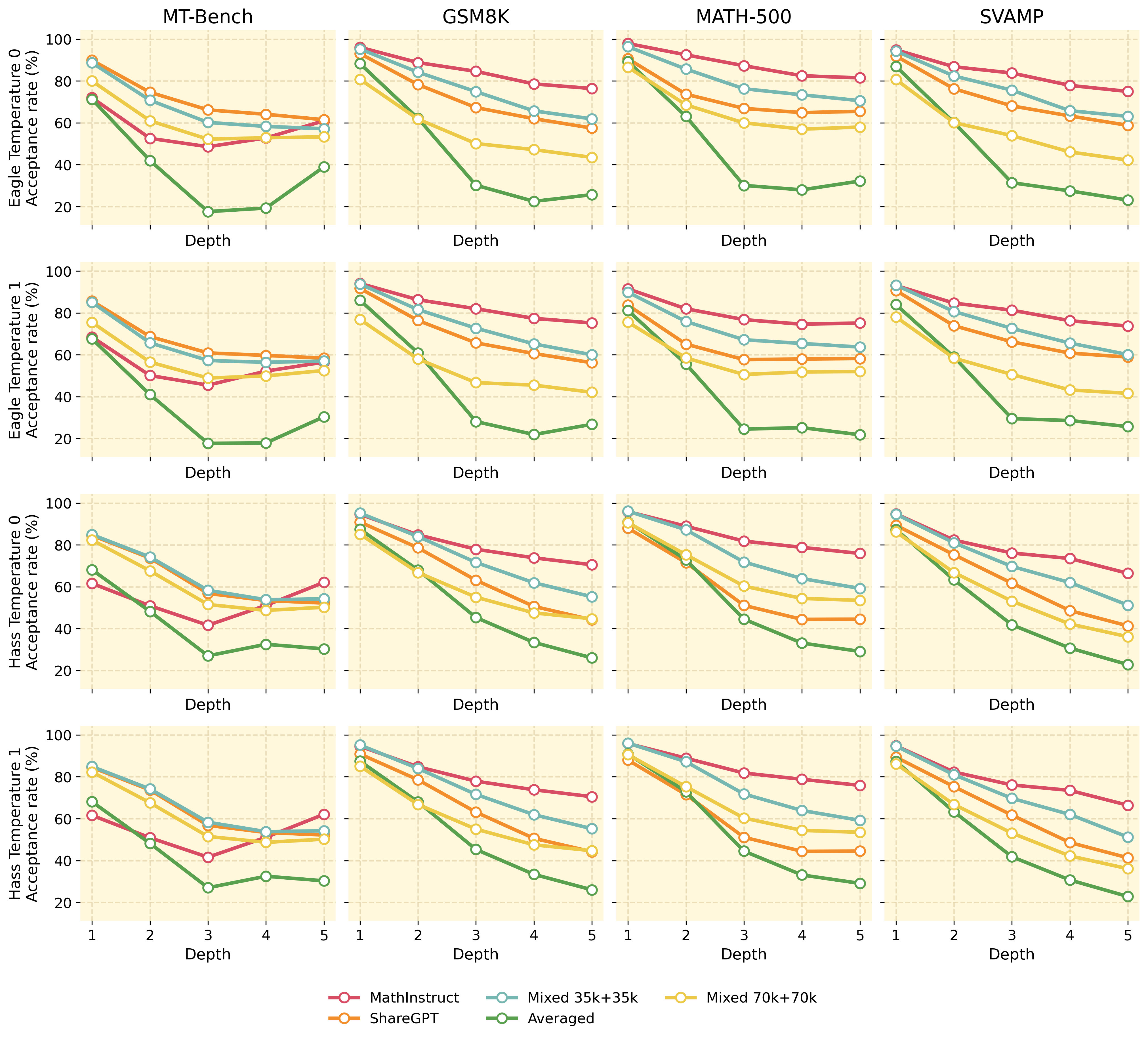}
    \caption{\textbf{Acceptance by Speculative Depth.} Acceptance rate is shown by draft depth for HASS and EAGLE-2 at temperatures $0$ and $1$ across MT-Bench, GSM8K, MATH-500, and SVAMP. Acceptance declines with depth for all variants, while domain specialization remains visible and often sharpens on reasoning-heavy tasks.}
    \label{fig:depths}
\end{figure}

\paragraph{Setup} We use Figure~\ref{fig:depths} and the depth tables to compare how the main draft variants behave as speculative depth increases across benchmarks.

\paragraph{Answer} The results show a clear depth effect. At shallow depths, mixed data drafts often perform best, suggesting an exploration benefit where broader proposal coverage increases the chance of producing acceptable early branches. As depth increases, the task matched specialist becomes more dominant, especially on reasoning benchmarks. This suggests exploitation, where deeper acceptance depends on sustained agreement between the drafter and the verifier. The composition results follow the same pattern because merged trees perform best by preserving diversity across specialists, while confidence routing helps when the system must choose a single drafter.

\paragraph{Takeaway} Speculative decoding appears to be both task aware and depth aware. Early proposal steps benefit more from coverage, while deeper accepted paths increasingly favor the better matched specialist.

%% file: sections/conclusion.tex
\section*{Discussion and Limitations}

The experiments support a simple but practically important conclusion: speculative decoding depends not only on the drafting backbone, but also on the relationship between the draft training distribution and the target workload. Once that dependency appears, the draft model becomes a systems choice rather than a fixed auxiliary component. A verifier paired with a mismatched draft is not merely weaker in the abstract; it is predictably weaker on specific task families.

This perspective also changes how multiple draft models should be used. Mixed-data training is a reasonable way to broaden coverage, but it does not eliminate the underlying specialization. When separate specialists are already available, keeping them separate and composing them at inference time is much more effective than collapsing them into one averaged checkpoint. The weight-space baseline is useful precisely because it fails: it shows that the relevant behavior is not preserved by naive interpolation.

Compared to the strongest single checkpoint, confidence routing reduces average speedup by $0.32\times$ and $0.35\times$ under EAGLE-2 and by $0.40\times$ and $0.47\times$ under HASS at temperatures $0$ and $1$, while merged-tree verification incurs a larger drop of $0.59\times$ and $0.62\times$ under EAGLE-2 and $0.72\times$ and $0.78\times$ under HASS. However, in a deployment setting that must serve two distinct task families, this overhead may be partly or fully offset when the best single checkpoint is weak on one of the tasks, since combining specialists can recover acceptance that a single drafter would lose.

At the same time, the present evidence is narrower than the strongest version of the claim. We evaluate one target model, two source domains, two speculative backbones, and four benchmarks. Acceptance length is our primary metric, so this paper does not establish end-to-end deployment trade-offs for routing or merged-tree verification. The routing policy is intentionally simple and confidence-based rather than learned or cost-aware. These limitations matter because they define the boundary of what the paper can currently claim.

\section*{Conclusion}

We asked whether speculative decoding improves when the drafter is trained for the downstream task and how multiple specialized drafters should be combined. The answer to the first question is yes: task-specific training produces clear domain specialization. The answer to the second is that inference-time composition is substantially stronger than weight-space averaging. Across both HASS and EAGLE-2, confidence routing improves over single-domain baselines and merged-tree verification achieves the highest acceptance length overall. More broadly, proposal quality in speculative decoding should be studied as a function of both draft architecture and draft training distribution, not architecture alone.

%% file: sections/appendix.tex
\clearpage
\section{Appendix}

This appendix collects the supporting entropy tables, the tree-merging utility used for merged-tree verification, and the depth-wise acceptance tables that complement Figure~\ref{fig:depths}. The goal is to make the evidence behind the main-text claims easy to audit without interrupting the main narrative.

\subsection{EAGLE-2 Entropy (Temperature 0)}
\begin{table}[H]
\centering
\resizebox{\textwidth}{!}{
\begingroup
\arrayrulecolor{black}
\rowcolors{3}{Cornsilk}{white}
\renewcommand{\arraystretch}{1.05}
\begin{tabular}{l l c c c c c c}
\toprule
Checkpoint & Benchmark & Draft Accepted & Draft Rejected & $\Delta$ Draft & Verifier Accepted & Verifier Rejected & $\Delta$ Verifier \\
\midrule
Averaged & GSM8K & 8.0257 & 9.2397 & $+1.2141$ & 0.1567 & 0.2698 & $+0.1131$ \\
Averaged & MATH-500 & 8.8405 & 9.8344 & $+0.9938$ & 0.2013 & 0.3787 & $+0.1774$ \\
Averaged & MT-Bench & 8.5128 & 9.4886 & $+0.9758$ & 0.2427 & 0.5796 & $+0.3368$ \\
Averaged & SVAMP & 7.8233 & 9.0891 & $+1.2658$ & 0.1703 & 0.2833 & $+0.1131$ \\
MathInstruct & GSM8K & 0.5284 & 1.0756 & $+0.5473$ & 0.1500 & 0.4246 & $+0.2746$ \\
MathInstruct & MATH-500 & 0.4567 & 0.8555 & $+0.3988$ & 0.1984 & 0.5067 & $+0.3083$ \\
MathInstruct & MT-Bench & 2.3867 & 3.8516 & $+1.4649$ & 0.2341 & 0.6212 & $+0.3871$ \\
MathInstruct & SVAMP & 0.5928 & 1.1607 & $+0.5679$ & 0.1639 & 0.4111 & $+0.2473$ \\
Mixed & GSM8K & 0.7074 & 1.2153 & $+0.5079$ & 0.1525 & 0.4671 & $+0.3146$ \\
Mixed & MATH-500 & 0.6302 & 1.0409 & $+0.4107$ & 0.1925 & 0.5430 & $+0.3505$ \\
Mixed & MT-Bench & 1.1839 & 1.8407 & $+0.6568$ & 0.2561 & 0.6736 & $+0.4175$ \\
Mixed & SVAMP & 0.8162 & 1.3367 & $+0.5205$ & 0.1717 & 0.4621 & $+0.2904$ \\
ShareGPT & GSM8K & 0.7434 & 1.0952 & $+0.3518$ & 0.1500 & 0.5075 & $+0.3574$ \\
ShareGPT & MATH-500 & 0.8926 & 1.2558 & $+0.3632$ & 0.1898 & 0.5766 & $+0.3868$ \\
ShareGPT & MT-Bench & 1.0404 & 1.6292 & $+0.5887$ & 0.2539 & 0.6600 & $+0.4061$ \\
ShareGPT & SVAMP & 0.8554 & 1.2391 & $+0.3837$ & 0.1667 & 0.5006 & $+0.3339$ \\
\bottomrule
\end{tabular}
\endgroup
}
\caption{\textbf{EAGLE-2 Entropy at Temperature 0.} Each row reports accepted-token and rejected-token entropy for one benchmark. Positive $\Delta$ means higher entropy for rejected tokens.}
\label{tab:entropy_eagle2_combined}
\end{table}

\subsection{HASS Entropy (Temperature 0)}
\begin{table}[H]
\centering
\resizebox{\textwidth}{!}{
\begingroup
\arrayrulecolor{black}
\rowcolors{3}{Cornsilk}{white}
\renewcommand{\arraystretch}{1.05}
\begin{tabular}{l l c c c c c c}
\toprule
Checkpoint & Benchmark & Draft Accepted & Draft Rejected & $\Delta$ Draft & Verifier Accepted & Verifier Rejected & $\Delta$ Verifier \\
\midrule
MathInstruct & GSM8K & 1.0731 & 1.4475 & $+0.3744$ & 0.1613 & 0.5508 & $+0.3895$ \\
MathInstruct & MATH-500 & 2.0806 & 2.1836 & $+0.1030$ & 0.1928 & 0.6625 & $+0.4698$ \\
MathInstruct & MT-Bench & 4.4777 & 5.2238 & $+0.7461$ & 0.2358 & 1.3139 & $+1.0780$ \\
MathInstruct & SVAMP & 1.3927 & 1.6778 & $+0.2851$ & 0.1765 & 0.5660 & $+0.3896$ \\
ShareGPT & GSM8K & 1.0690 & 1.7551 & $+0.6861$ & 0.1549 & 0.5937 & $+0.4388$ \\
ShareGPT & MATH-500 & 2.2460 & 2.3454 & $+0.0995$ & 0.1814 & 0.7575 & $+0.5761$ \\
ShareGPT & MT-Bench & 1.9162 & 2.2477 & $+0.3315$ & 0.2796 & 0.8882 & $+0.6086$ \\
ShareGPT & SVAMP & 1.3445 & 2.0791 & $+0.7345$ & 0.1695 & 0.6139 & $+0.4444$ \\
Mixed & GSM8K & 1.2369 & 1.7589 & $+0.5220$ & 0.1597 & 0.5287 & $+0.3690$ \\
Mixed & MATH-500 & 1.9626 & 2.2128 & $+0.2502$ & 0.1876 & 0.6563 & $+0.4687$ \\
Mixed & MT-Bench & 2.0482 & 2.4916 & $+0.4433$ & 0.2766 & 0.9013 & $+0.6247$ \\
Mixed & SVAMP & 1.3306 & 1.8352 & $+0.5047$ & 0.1742 & 0.5501 & $+0.3759$ \\
Averaged & GSM8K & 2.6208 & 3.2743 & $+0.6535$ & 0.1610 & 0.8306 & $+0.6696$ \\
Averaged & MATH-500 & 3.8384 & 3.3019 & $-0.5364$ & 0.1786 & 0.9489 & $+0.7703$ \\
Averaged & MT-Bench & 4.2061 & 3.5950 & $-0.6110$ & 0.2362 & 1.4447 & $+1.2085$ \\
Averaged & SVAMP & 2.5409 & 3.1844 & $+0.6434$ & 0.1739 & 0.8523 & $+0.6784$ \\
\bottomrule
\end{tabular}
\endgroup
}
\caption{\textbf{HASS Entropy at Temperature 0.} Each row reports accepted-token and rejected-token entropy for one benchmark. Positive $\Delta$ means higher entropy for rejected tokens.}
\label{tab:entropy_hass_combined}
\end{table}

\subsection{Tree Merge Utility}
\begin{tcolorbox}[colback=Gray, colframe=black, title={\texttt{\_merge\_trees}}, left=4mm, right=4mm, top=3mm, bottom=3mm]
\footnotesize
\begin{verbatim}
def _merge_trees(
    draft_tokens1, retrieve_indices1, tree_mask1, tree_pos1,
    draft_tokens2, retrieve_indices2, tree_mask2, tree_pos2,
):
    n1 = draft_tokens1.shape[1] - 1
    n2 = draft_tokens2.shape[1] - 1
    N = n1 + n2 + 1
    device = draft_tokens1.device
    dtype = tree_mask1.dtype

    merged_draft = torch.cat([draft_tokens1, draft_tokens2[0, 1:][None]], dim=1)

    merged_mask = torch.zeros(N, N, device=device, dtype=dtype)
    merged_mask[0, 0] = 1.0
    merged_mask[1:n1 + 1, :n1 + 1] = tree_mask1[0, 0, 1:, :]
    merged_mask[n1 + 1:, 0] = 1.0
    merged_mask[n1 + 1:, n1 + 1:] = tree_mask2[0, 0, 1:, 1:]
    merged_mask = merged_mask[None, None]

    merged_pos = torch.cat([tree_pos1, tree_pos2[1:]])

    d1, d2 = retrieve_indices1.shape[1], retrieve_indices2.shape[1]
    max_d = max(d1, d2)
    ri1 = F.pad(retrieve_indices1, (0, max_d - d1), value=-1)
    ri2 = retrieve_indices2.clone()
    ri2[ri2 > 0] += n1
    ri2 = F.pad(ri2, (0, max_d - d2), value=-1)
    merged_retrieve = torch.cat([ri1, ri2], dim=0)

    return merged_draft, merged_retrieve, merged_mask, merged_pos
\end{verbatim}
\end{tcolorbox}

\subsection{Correctness of routing and merged-tree verification}
Let $Q(\cdot \mid y_{1:t})$ denote the target model’s continuation distribution from prefix $y_{1:t}$.
For any (packed) draft tree $\mathcal{T}$ rooted at $y_{1:t}$, let $\mathrm{Dec}(y_{1:t};\mathcal{T})$
denote the random continuation produced by running one verifier call on $\mathcal{T}$ and then continuing
with the standard speculative procedure.

\paragraph{Assumption A.1 (Lossless verification for a fixed valid tree).}
A packed tree $\mathcal{T}$ is \emph{valid} if the verifier pass on $\mathcal{T}$ produces, for every node,
the same target-side conditionals $q(\cdot \mid \text{its path-prefix})$ that the target model would produce
under standalone autoregressive evaluation along that node’s path.
For every valid $\mathcal{T}$ and every measurable set of continuations $B$,
\[
\Pr\!\big(\mathrm{Dec}(y_{1:t};\mathcal{T}) \in B \,\big|\, y_{1:t}, \mathcal{T}\big) \;=\; Q(B \mid y_{1:t}).
\]
(This is the standard lossless speculative-decoding guarantee used throughout the paper.)

\paragraph{Lemma A.1 (Mixtures over valid trees remain lossless).}
Let $\mathcal{T}$ be any \emph{random} valid tree (possibly generated by any draft model(s)).
Then for every set $B$,
\[
\Pr\!\big(\mathrm{Dec}(y_{1:t};\mathcal{T}) \in B \,\big|\, y_{1:t}\big) \;=\; Q(B \mid y_{1:t}).
\]
\emph{Proof.} By the tower property,
\[
\Pr(\mathrm{Dec}\in B \mid y_{1:t})
=\mathbb{E}\!\left[\Pr(\mathrm{Dec}\in B \mid y_{1:t},\mathcal{T}) \mid y_{1:t}\right]
=\mathbb{E}\!\left[Q(B\mid y_{1:t}) \mid y_{1:t}\right]
=Q(B\mid y_{1:t}). \qedhere
\]

\paragraph{Proposition A.1 (Correctness of routing).}
Let $\mathcal{T}_{\text{math}}$ and $\mathcal{T}_{\text{chat}}$ be two valid draft trees generated from the same prefix
$y_{1:t}$. Let $g$ be any (possibly randomized) routing rule that depends only on draft-side quantities available
\emph{before} verification (e.g., confidences/entropies/tree statistics), and define the selected tree
$\mathcal{T}^\star = \mathcal{T}_{g(y_{1:t},\mathcal{T}_{\text{math}},\mathcal{T}_{\text{chat}})}$.
Then routing is distribution-preserving:
\[
\Pr\!\big(\mathrm{Dec}(y_{1:t};\mathcal{T}^\star)\in B \,\big|\, y_{1:t}\big) \;=\; Q(B\mid y_{1:t})
\quad \text{for all } B.
\]
\emph{Proof.} $\mathcal{T}^\star$ is a random valid tree (a draft-side function of $(\mathcal{T}_{\text{math}},\mathcal{T}_{\text{chat}})$),
so the claim follows immediately from Lemma~A.1. \qed

\paragraph{Lemma A.2 (Verifier invariance under masked concatenation).}
Let the verifier be any transformer-style model that computes per-token logits from (token ids, position ids, attention mask).
Consider two packed verifier inputs $(X,M,P)$ and $(X',M',P')$ with a shared index set $S$ such that:
(i) $X|_S = X'|_S$ and $P|_S = P'|_S$; 
(ii) $M|_{S\times S} = M'|_{S\times S}$; and
(iii) tokens in $S$ do not attend outside $S$ in either input, i.e. for all $i\in S$ and $j\notin S$,
$M_{ij}=M'_{ij}=0$.
Then the verifier logits on indices in $S$ are identical under the two packed inputs.
\emph{Proof.} Induct over transformer layers. At layer $0$, hidden states on $S$ match because token embeddings and position encodings match.
Assume hidden states on $S$ match at layer $\ell-1$. At layer $\ell$, each token $i\in S$ attends only to tokens $j$ with $M_{ij}=1$,
and by (iii) all such $j$ lie in $S$. By (ii) the mask on $S\times S$ matches, and by the inductive hypothesis the keys/values of all
visible $j\in S$ match. Therefore the attention output for each $i\in S$ matches; the remaining sublayers are pointwise with shared parameters,
so hidden states on $S$ match at layer $\ell$. Hence the final logits on $S$ match. \qed

\paragraph{Proposition A.2 (Correctness of merged-tree verification).}
Let $\mathcal{T}_{\text{math}}$ and $\mathcal{T}_{\text{chat}}$ be valid trees from prefix $y_{1:t}$, each with its own packed
representation (tokens, tree attention mask, and depth-based position ids) used for standalone tree verification.
Construct the merged tree $\mathcal{T}_{\cup}$ by (a) sharing the root, (b) concatenating the non-root nodes of both trees, (c) using
an attention mask that preserves each subtree’s ancestry relations and \emph{masks all cross-subtree attention}, and (d) assigning each
node the same depth-based position id it had in its source tree.

Then (i) every node in the merged verifier pass receives exactly the same target-side conditional distribution as in standalone verification
of its source subtree, and consequently (ii) merged-tree verification is distribution-preserving:
\[
\Pr\!\big(\mathrm{Dec}(y_{1:t};\mathcal{T}_{\cup})\in B \,\big|\, y_{1:t}\big) \;=\; Q(B\mid y_{1:t})
\quad \text{for all } B.
\]

\emph{Proof.}
Fix $s\in\{\text{math},\text{chat}\}$ and let $S_s$ denote the index set of the shared root together with all nodes coming from subtree $s$
inside the merged packing.
By construction, the merged packed input agrees with the standalone packed input on $S_s$ (tokens, depth-based positions,
and within-subtree attention), and nodes in $S_s$ do not attend to nodes outside $S_s$ because all cross-subtree attention is masked.
Therefore, by Lemma~A.2, the verifier logits (hence $q(\cdot\mid\cdot)$) on all nodes in subtree $s$ are identical to standalone verification.
This holds for both subtrees, so $\mathcal{T}_{\cup}$ is a valid tree in the sense of Assumption~A.1.

Applying Assumption~A.1 to the fixed valid tree $\mathcal{T}_{\cup}$ yields
$\Pr(\mathrm{Dec}(y_{1:t};\mathcal{T}_{\cup})\in B \mid y_{1:t},\mathcal{T}_{\cup})=Q(B\mid y_{1:t})$ for all $B$.
Unconditioning (or equivalently applying Lemma~A.1) gives the claimed distribution preservation. \qed

\paragraph{Corollary A.1.}
Both routing (Proposition~A.1) and merged-tree verification (Proposition~A.2) preserve the target-model output distribution.
They may change proposal quality, acceptance length, and runtime, but not the verifier’s sampling law.

\subsection{EAGLE-2 Acceptance Rates by Depth}
\subsubsection{Temperature 0}
\begin{table}[H]
\centering
\resizebox{0.8\textwidth}{!}{
\begingroup
\arrayrulecolor{black}
\rowcolors{3}{Cornsilk}{white}
\renewcommand{\arraystretch}{1.05}
\begin{tabular}{c c l c c c c}
\toprule
Depth & Variant & MT-Bench & GSM8K & MATH-500 & SVAMP & Avg \\
\midrule
1 & MathInstruct & 72.0\% & 96.1\% & 97.9\% & 94.9\% & 90.2\% \\
1 & ShareGPT & 89.9\% & 93.1\% & 90.6\% & 91.9\% & 91.4\% \\
1 & Mixed 35k+35k & 88.7\% & 95.3\% & 96.4\% & 94.3\% & 93.7\% \\
1 & Averaged & 71.3\% & 88.3\% & 89.3\% & 87.0\% & 84.0\% \\
1 & Mixed 70k+70k & 80.0\% & 80.8\% & 86.6\% & 80.7\% & 82.0\% \\
2 & MathInstruct & 52.6\% & 88.8\% & 92.5\% & 86.8\% & 80.2\% \\
2 & ShareGPT & 74.6\% & 78.3\% & 73.7\% & 76.3\% & 75.7\% \\
2 & Mixed 35k+35k & 70.8\% & 84.2\% & 85.7\% & 82.4\% & 80.8\% \\
2 & Averaged & 42.0\% & 62.2\% & 63.1\% & 60.2\% & 56.9\% \\
2 & Mixed 70k+70k & 61.0\% & 61.8\% & 68.6\% & 60.1\% & 62.9\% \\
3 & MathInstruct & 48.6\% & 84.6\% & 87.3\% & 83.8\% & 76.1\% \\
3 & ShareGPT & 66.2\% & 67.3\% & 66.9\% & 68.1\% & 67.1\% \\
3 & Mixed 35k+35k & 60.2\% & 74.9\% & 76.3\% & 75.6\% & 71.8\% \\
3 & Averaged & 17.6\% & 30.2\% & 30.1\% & 31.4\% & 27.3\% \\
3 & Mixed 70k+70k & 52.2\% & 50.1\% & 60.0\% & 53.9\% & 54.0\% \\
4 & MathInstruct & 52.8\% & 78.6\% & 82.5\% & 77.9\% & 72.9\% \\
4 & ShareGPT & 64.1\% & 62.0\% & 64.9\% & 63.3\% & 63.6\% \\
4 & Mixed 35k+35k & 58.3\% & 65.7\% & 73.4\% & 65.8\% & 65.8\% \\
4 & Averaged & 19.3\% & 22.5\% & 28.0\% & 27.5\% & 24.3\% \\
4 & Mixed 70k+70k & 52.9\% & 47.2\% & 57.0\% & 46.1\% & 50.8\% \\
5 & MathInstruct & 61.0\% & 76.4\% & 81.5\% & 75.0\% & 73.5\% \\
5 & ShareGPT & 61.5\% & 57.5\% & 65.5\% & 58.8\% & 60.8\% \\
5 & Mixed 35k+35k & 57.2\% & 61.9\% & 70.6\% & 63.1\% & 63.2\% \\
5 & Averaged & 39.0\% & 25.7\% & 32.2\% & 23.1\% & 30.0\% \\
5 & Mixed 70k+70k & 53.3\% & 43.5\% & 58.0\% & 42.2\% & 49.3\% \\
\bottomrule
\end{tabular}
\endgroup
}
\caption{\textbf{EAGLE-2 Acceptance by Depth at Temperature 0.} Higher rows correspond to shallower draft positions in the speculative tree.}
\label{tab:eagle_depth}
\end{table}

\subsubsection{Temperature 1}
\begin{table}[H]
\centering
\resizebox{0.8\textwidth}{!}{
\begingroup
\arrayrulecolor{black}
\rowcolors{3}{Cornsilk}{white}
\renewcommand{\arraystretch}{1.05}
\begin{tabular}{c c l c c c c}
\toprule
Depth & Variant & MT-Bench & GSM8K & MATH-500 & SVAMP & Avg \\
\midrule
1 & MathInstruct & 68.5\% & 94.1\% & 91.5\% & 93.1\% & 86.8\% \\
1 & ShareGPT & 85.7\% & 91.7\% & 83.7\% & 90.7\% & 88.0\% \\
1 & Mixed 35k+35k & 85.1\% & 93.8\% & 89.9\% & 93.3\% & 90.5\% \\
1 & Averaged & 67.6\% & 86.1\% & 81.2\% & 84.0\% & 79.7\% \\
1 & Mixed 70k+70k & 75.5\% & 76.9\% & 75.6\% & 78.1\% & 76.5\% \\
2 & MathInstruct & 50.1\% & 86.3\% & 82.0\% & 84.7\% & 75.8\% \\
2 & ShareGPT & 68.7\% & 76.4\% & 65.1\% & 73.9\% & 71.0\% \\
2 & Mixed 35k+35k & 65.7\% & 81.6\% & 75.9\% & 80.7\% & 76.0\% \\
2 & Averaged & 41.0\% & 60.9\% & 55.6\% & 59.0\% & 54.1\% \\
2 & Mixed 70k+70k & 56.5\% & 58.0\% & 58.5\% & 58.4\% & 57.8\% \\
3 & MathInstruct & 45.5\% & 82.0\% & 76.8\% & 81.3\% & 71.4\% \\
3 & ShareGPT & 60.9\% & 65.7\% & 57.7\% & 66.2\% & 62.6\% \\
3 & Mixed 35k+35k & 57.3\% & 72.7\% & 67.2\% & 72.6\% & 67.5\% \\
3 & Averaged & 17.7\% & 28.1\% & 24.5\% & 29.5\% & 24.9\% \\
3 & Mixed 70k+70k & 48.9\% & 46.7\% & 50.6\% & 50.6\% & 49.2\% \\
4 & MathInstruct & 52.2\% & 77.4\% & 74.6\% & 76.3\% & 70.1\% \\
4 & ShareGPT & 59.7\% & 60.6\% & 58.0\% & 60.8\% & 59.8\% \\
4 & Mixed 35k+35k & 56.4\% & 65.2\% & 65.4\% & 65.6\% & 63.1\% \\
4 & Averaged & 17.9\% & 21.9\% & 25.2\% & 28.6\% & 23.4\% \\
4 & Mixed 70k+70k & 49.9\% & 45.5\% & 51.8\% & 43.2\% & 47.6\% \\
5 & MathInstruct & 56.5\% & 75.2\% & 75.2\% & 73.7\% & 70.1\% \\
5 & ShareGPT & 58.4\% & 56.2\% & 58.2\% & 58.9\% & 57.9\% \\
5 & Mixed 35k+35k & 57.0\% & 60.0\% & 63.7\% & 60.1\% & 60.2\% \\
5 & Averaged & 30.4\% & 26.8\% & 21.8\% & 25.7\% & 26.2\% \\
5 & Mixed 70k+70k & 52.5\% & 42.1\% & 52.0\% & 41.6\% & 47.0\% \\
\bottomrule
\end{tabular}
\endgroup
}
\caption{\textbf{EAGLE-2 Acceptance by Depth at Temperature 1.} Higher rows correspond to shallower draft positions in the speculative tree.}
\label{tab:eagle_depth_temp1}
\end{table}

\subsection{HASS Acceptance Rates by Depth}
\subsubsection{Temperature 0}
\begin{table}[H]
\centering
\resizebox{0.8\textwidth}{!}{
\begingroup
\arrayrulecolor{black}
\rowcolors{3}{Cornsilk}{white}
\renewcommand{\arraystretch}{1.05}
\begin{tabular}{c c l c c c c}
\toprule
Depth & Variant & MT-Bench & GSM8K & MATH-500 & SVAMP & Avg \\
\midrule
1 & MathInstruct & 61.6\% & 94.6\% & 95.9\% & 94.8\% & 86.7\% \\
1 & ShareGPT & 84.5\% & 90.9\% & 88.0\% & 89.4\% & 88.2\% \\
1 & Mixed 35k+35k & 84.9\% & 95.2\% & 96.1\% & 94.7\% & 92.7\% \\
1 & Averaged & 68.1\% & 87.5\% & 90.9\% & 87.4\% & 83.5\% \\
1 & Mixed 70k+70k & 82.3\% & 85.1\% & 90.6\% & 86.2\% & 86.0\% \\
2 & MathInstruct & 50.9\% & 84.8\% & 88.9\% & 82.3\% & 76.7\% \\
2 & ShareGPT & 73.7\% & 78.6\% & 71.5\% & 75.3\% & 74.8\% \\
2 & Mixed 35k+35k & 74.2\% & 84.0\% & 87.2\% & 81.0\% & 81.6\% \\
2 & Averaged & 48.2\% & 68.0\% & 72.9\% & 63.3\% & 63.1\% \\
2 & Mixed 70k+70k & 67.6\% & 66.8\% & 75.3\% & 66.8\% & 69.1\% \\
3 & MathInstruct & 41.6\% & 77.9\% & 81.8\% & 76.1\% & 69.4\% \\
3 & ShareGPT & 56.8\% & 63.1\% & 51.1\% & 61.7\% & 58.2\% \\
3 & Mixed 35k+35k & 58.3\% & 71.6\% & 71.8\% & 69.7\% & 67.9\% \\
3 & Averaged & 27.0\% & 45.3\% & 44.5\% & 41.8\% & 39.6\% \\
3 & Mixed 70k+70k & 51.5\% & 55.0\% & 60.3\% & 53.1\% & 55.0\% \\
4 & MathInstruct & 51.1\% & 73.8\% & 78.8\% & 73.5\% & 69.3\% \\
4 & ShareGPT & 53.4\% & 50.6\% & 44.4\% & 48.6\% & 49.2\% \\
4 & Mixed 35k+35k & 53.8\% & 61.9\% & 63.9\% & 62.0\% & 60.4\% \\
4 & Averaged & 32.5\% & 33.4\% & 33.1\% & 30.7\% & 32.4\% \\
4 & Mixed 70k+70k & 48.7\% & 47.5\% & 54.4\% & 42.2\% & 48.2\% \\
5 & MathInstruct & 62.1\% & 70.5\% & 75.9\% & 66.4\% & 68.7\% \\
5 & ShareGPT & 52.2\% & 44.2\% & 44.5\% & 41.3\% & 45.5\% \\
5 & Mixed 35k+35k & 54.2\% & 55.2\% & 59.2\% & 51.2\% & 54.9\% \\
5 & Averaged & 30.3\% & 26.0\% & 29.1\% & 22.8\% & 27.0\% \\
5 & Mixed 70k+70k & 50.2\% & 44.7\% & 53.5\% & 36.1\% & 46.1\% \\
\bottomrule
\end{tabular}
\endgroup
}
\caption{\textbf{HASS Acceptance by Depth at Temperature 0.} Higher rows correspond to shallower draft positions in the speculative tree.}
\label{tab:hass_depth_temp0}
\end{table}